\documentclass{article}
\usepackage{url}
\usepackage{listings}
\usepackage{multicol}
\usepackage{authblk}
\usepackage{multirow}
\usepackage{graphicx}
\usepackage{hyperref}
\usepackage{fullpage}
\usepackage{color}
\usepackage{mathtools}
\usepackage{booktabs}       
\usepackage{algorithm}
\usepackage{algpseudocode}
\usepackage[T1]{fontenc}
\usepackage{wrapfig}

\providecommand{\keywords}[1]
{
  \small	
  \textbf{\textit{Keywords---}} #1
}

\begin{document}

\title{Parser Extraction of Triples in Unstructured Text}
\author{Shaun D'Souza \\
Technical Lead, Wipro Limited, Bangalore, Karnataka, India \\
shaun.dsouza1@wipro.com}

\date{}

\maketitle

\begin{abstract}
The web contains vast repositories of unstructured text. We investigate the opportunity for building a knowledge graph from these text sources. We generate a set of triples which can be used in knowledge gathering and integration. We define the architecture of a language compiler for processing subject-predicate-object triples using the OpenNLP parser. We implement a depth-first search traversal on the POS tagged syntactic tree appending predicate and object information. A parser enables higher precision and higher recall extractions of syntactic relationships across conjunction boundaries. We are able to extract 2-2.5 times the correct extractions of ReVerb. The extractions are used in a variety of semantic web applications and question answering. We verify extraction of 50,000 triples on the ClueWeb dataset.
\end{abstract}

\keywords{Open information extraction, Relation extraction, NLP}

\section{Introduction}

There is a considerable amount of research in natural language processing (NLP). With the availability of a larger set of NLP tools like OpenNLP~\cite{opennlp2011apache}, it is today possible to POS tag and chunk vast amount of unstructured text that is available on the internet. Projects like ClueWeb, OpenIE and Wikipedia provide a corpus of text data which can be used for ontological engineering. OpenNLP supports the POS tagging and chunking of data. It outputs a parse tree for the data which encapsulates the syntactic content in a n-ary tree data structure. POS tag data provides a higher level of understanding as compared to a bag of words approach to web search today. We explore opportunities for language inference and understanding through subject-predicate-object analysis of web scale unstructured data.

Various methods are used to extract subject-predicate-object triples in unstructured data. DBpedia extractor is used to generate triples using annotated field information in Wikipedia. OpenIE~\cite{etzioni2011open} used POS and chunker data while ClauseIE~\cite{del2013clausie} uses a parser to output a set of word triples.

Bootstrapping functions use N-gram models to generate a template for a given combination of noun phrases. These are used to search a larger corpus of data for similar templates and generate values. NER taggers are used to annotate person and location information.
We assume a context free grammar (CFG) for English language~\cite{hopcroft2001introduction}.

\begin{eqnarray*}
G & = & (N, \Sigma, R, S) \\
\\
N & \in & \{non-terminal\ symbols\} \\
\Sigma & \in & \{terminal\ symbols\} \\
R & \in & \{rules\}\ of\ the\ form\ X \rightarrow Y_1Y_n\ for\ n \geq 0, X \in N, Y_i \in (N \cup \Sigma) \\
S & \in & N\ start\ symbol\ \{TOP\} \\
\\
N & = & \{S, NP, VP, PP, DT, VB, NN, IN\} \\
S & = & S \\
\Sigma & = & word\ in\ the\ English\ language \\
\end{eqnarray*}

\begin{eqnarray*}
R & = & S \rightarrow NP\ VP \\
& & VP \rightarrow VB \\
& & VP \rightarrow VB\ NP \\
& & VP \rightarrow VP\ PP \\
& & NP \rightarrow DT\ NN \\
& & NP \rightarrow NP\ PP \\
& & PP \rightarrow IN\ NP \\
\end{eqnarray*}

\section{Research Method}
We found a limitation of extractors that were unable to extract the verb phrase accurately and instead appended a large amount of additional words including the trailing noun and preposition context. The extractors were unable to process sentence and conjunction values resulting in incorrect verb and object phrases. A parse tree is able to capture conjunction and object phrase information correctly. Although there is an overhead on the parsing time.

We evaluate the parser tree for sequences of NP noun phrases (subject, object) and VB - verbs (predicate). OpenNLP generates a parse tree using the CFG rules. We implement an in-order traversal of the syntactic tree to detect SVO phrases. We maintain a list of all NP phrases in the sentence. We then traverse the tree to detect subject object pairs and the predicate.

\begin{algorithm}
\begin{algorithmic}
\Function {SUBJECT-NOUN-PHRASE}{$parse$}
    \State $kids \gets CHILD(parse)$
    \For {i = 1 to SIZE(kids)}
		\If {TYPE(kids[i]) = NP}
			\State subject = kids[i]
			\For {j = i + 1 to SIZE(kids)}
				\If {TYPE(kids[j]) = VP {\bf or} PP {\bf or} SBAR}
					\State explored $\leftarrow$ an empty set 
					\While {kids[j] not in explored}
						\State extraction $\leftarrow$ APPEND(subject, PREDICATE-VERB-PHRASE(kids[j]))
						\State PRINT(extraction)
					\EndWhile
				\EndIf
			\EndFor
		\EndIf
		\State SUBJECT-NOUN-PHRASE(kids[i])
	\EndFor
\EndFunction
\\

\Function {PREDICATE-VERB-PHRASE}{$parse$} 
	\Return solution, failure
	\State $kids \gets CHILD(parse)$
	\State initialize predicate string to be empty

	\For {i = 1 to SIZE(kids)}
		\If {TYPE(kids[i]) = VP {\bf or} S}
			\If {kids[i] not in explored}
				\State \Return APPEND(predicate, PREDICATE-VERB-PHRASE(kids[i]))
			\EndIf
		\ElsIf {TYPE(kids[i]) = VB {\bf or} JJ {\bf or} RB {\bf or} MD {\bf or} TO {\bf or} ADVP {\bf or} DT {\bf or} NN {\bf or} IN}
			\State $predicate \gets APPEND(predicate, kids[i])$

			\For {j = i + 1 to SIZE(kids)}
				\If {TYPE(kids[j]) = NP {\bf or} PP {\bf or} ADJP {\bf or} S {\bf or} SBAR}
					\State \Return APPEND(predicate, OBJECT-NOUN-PHRASE(kids[j]))
				\EndIf
			\EndFor
		\EndIf
	\EndFor
	
	\State add parse to explored
	\State \Return failure
\EndFunction
\end{algorithmic}
\caption{Subject-predicate phrase algorithm}
\label{fig:subject}
\end{algorithm}

\begin{algorithm}
\begin{algorithmic}
\Function {OBJECT-NOUN-PHRASE}{$parse$}
	\Return solution, failure
	\State $found \gets false$
	\State $kids \gets CHILD(parse)$
	\State initialize object string to be empty

	\For {i = 1 to SIZE(kids)}
		\If {TYPE(kids[i]) = NP {\bf or} S}
			\State $found \gets true$
			\If {kids[i] not in explored}
				\State \Return APPEND(object, OBJECT-NOUN-PHRASE(kids[i]))
			\Else 
				\State \Return APPEND(object, GET-COVERED-TEXT(kids[i]))
			\EndIf
		\ElsIf {TYPE(kids[i]) = PP}
			\If {kids[i] not in explored}
				\State \Return APPEND(object, OBJECT-PREPOSTION-PHRASE(kids[i]))
			\Else
				\State \Return APPEND(object, GET-COVERED-TEXT(kids[i]))
			\EndIf
		\ElsIf {TYPE(kids[i]) = IN {\bf or} TO}
			\State $object \gets APPEND(object, kids[i])$
		\EndIf
	\EndFor
	
	\State add parse to explored
	\If {not found and TYPE(parse) = NP}
		\State \Return APPEND(object, parse)
	\EndIf

	\State \Return failure
\EndFunction
\\
\Function {OBJECT-PREPOSITION-PHRASE}{$parse$}
	\Return solution, failure
	\State $kids \gets CHILD(parse)$
	\State initialize preposition string to be empty

	\For {i = 1 to SIZE(kids)}
		\If {TYPE(kids[i]) = NP and not in explored}
			\State \Return APPEND(preposition, OBJECT-NOUN-PHRASE(kids[i]))
		\ElsIf {TYPE(kids[i]) = PP and not in explored}
			\State \Return APPEND(preposition, OBJECT-PREPOSTION-PHRASE(kids[i]))
		\ElsIf {TYPE(kids[i]) = IN {\bf or} TO {\bf or} JJ {\bf or} ADVP}
			\State $preposition \gets APPEND(preposition, kids[i])$
		\EndIf
	\EndFor
	
	\State add parse to explored
	\State \Return failure
\EndFunction
\end{algorithmic}
\caption{Object phrase algorithm}
\label{fig:object}
\end{algorithm}

We implement a depth-first search on the n-ary parse tree. We search the parse tree for a noun-verb phrase indicating the subject-predicate - Fig.~\ref{fig:subject}. The noun phrase is used as the subject in the clause. We look for a verb phrase VP or preposition phrase PP in the siblings. In the case of subsequent conjunctions CC and WHNP phrases, we continue to search the sibling nodes. For all found VP, PP we search for the predicate clause in the sentence. A predicate clause consists of a sequence of verb, adjectives, adverb and modal identifiers. These are appended to a string of predicates. VP phrases are searched recursively till we find a terminal NP object clause. We represent the SVO in the triples format. We use a training set of 200 phrases from earlier publications on information extraction. These give us a range of parse trees to evaluate the search on and refine.

Earlier work on information extraction was limited to the capabilities of the POS and Chunker tags. Verb phrases were detected using statistical probabilities of frequently occurring patterns in the English language. We implement a rigorous parse tree design which preserves the language syntax of the text data.

As there is a high availability of computing today in the cloud, we implement the SVO parser as an offline function to process the syntactic tree. We parse all the sentences in the text and generate a parsed output. This is subsequently used to generate the SVO triples. With the availability of computing we can improve performance of the parser by parallelizing the parsing of input sentences.

We contrast the SVO triples with past research including OpenIE and ClauseIE. We find that a parser based approach is able to extract a large number of SVO's accurately. Availability of a syntactic parse tree also enables us to extract triples with reduced ambiguity. The obtained triples map exactly to sub-trees in the sentence parse tree and capture all the semantic information - subject predicate. The n-ary parse tree encapsulates the syntactic structure of the sentence completely.

We are able to precisely extract SVO information. In the initial revision of the code we implemented predicate extractions to include the trailing noun phrase. This was updated to resolve the object clause to contain the noun phrase NP and a trailing preposition phrase PP - Fig.~\ref{fig:object}. We use a set of heuristics to maximize the number of triples generated for each noun phrase, verb phrase.

\section{Results and Analysis}

The SVO extractions are coherent as OpenNLP captures the language syntax in the parse tree. We compare the number of extractions with the ReVerb extractor. We observe a larger number of triples as we are searching for all noun phrases in the object. The NLP parser is able to extract a large number of triples matching ReVerb and ClausIE.

Example sentence \\
The principal opposition parties boycotted the polls after accusations of vote rigging, and the only other name on the ballot was a little known challenger from a marginal political party

\begin{figure}[!h]
    \centering
	\caption{An example sentence parse tree.}
	\includegraphics[width=\textwidth]{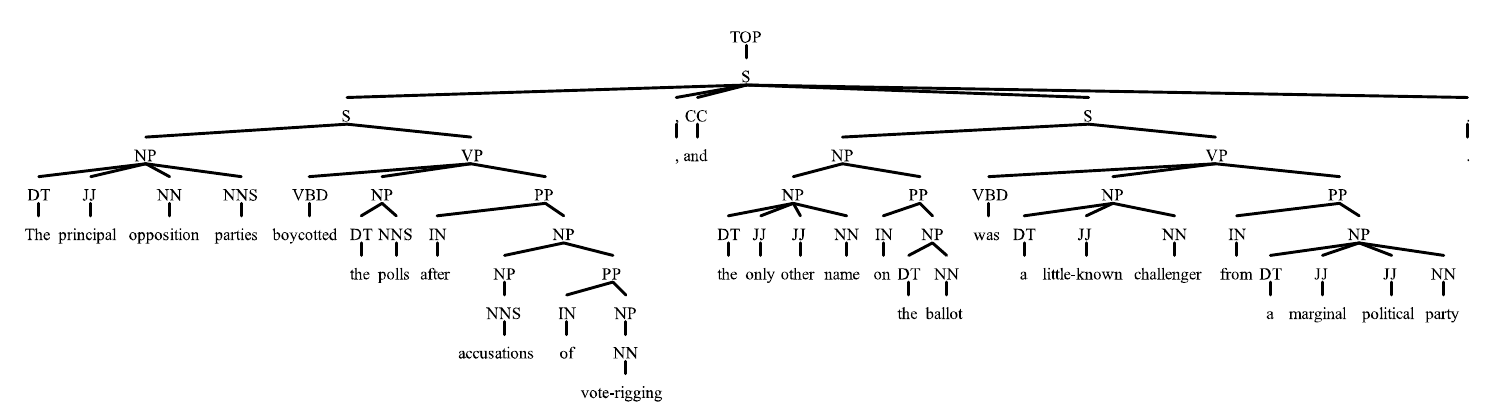}
	\label{fig:parse}
\end{figure}

(``The principal opposition parties'', ``boycotted'', ``the polls'') \\
(``The principal opposition parties'', ``boycotted'', ``the polls after accusations'') \\
(``The principal opposition parties'', ``boycotted'', ``the polls after accusations of vote rigging'') \\
(``the only other name on the ballot'', ``was'', ``a little known challenger'') \\
(``the only other name on the ballot'', ``was'', ``a little known challenger from a marginal political party'') \\

The above extractions are labelled correctly in the ReVerb dataset and contain some redundant extractions. We evaluated the parser extraction on the ClueWeb12 dataset and were able to extract more than 50,000 triples. We found that the parser was able to perform on par with ReVerb and ClausIE. This was achieved using the syntactic functionality in the parse tree - Fig.~\ref{fig:parse}. It demonstrates the ability of a parser based approach in extracting high quality triples.  

We verified the extractions for a sample set of sentences in the OpenIE and ClausIE publications. These were used to ensure precision in the parser extractions. We additionally ran the parser on the ClueWeb data and compared the number of extractions with the alternative approaches. We measured the distribution of the noun and verb sub-trees in the sentence text - Table ~\ref{table:phrase}. We found that 10\% of the phrases were prepositional. The density of the noun and verb phrases are in agreement with the English context free grammar (CFG).

\begin{table}[!h]
	\caption{Phrase distribution}
	\label{table:phrase}
	\centering
	\begin{tabular}{ lr }
	\toprule
	Noun & Frequency \\
	\midrule
	NP $\rightarrow$ NN	& 14\% \\
	NP $\rightarrow$ NP PP & 12\% \\
	NP $\rightarrow$ DT NN & 12\% \\
	NP $\rightarrow$ NN NN & 6\% \\
	\bottomrule
    \end{tabular}
	\quad
	\begin{tabular}{ lr }
	\toprule
	Verb & Frequency \\
	\midrule
	VP $\rightarrow$ VB NP & 16\% \\
	VP $\rightarrow$ VB VP & 10\% \\
	VP $\rightarrow$ TO VP & 9\% \\
	VP $\rightarrow$ VB PP & 8\% \\
	VP $\rightarrow$ VB & 6\% \\
	\bottomrule
    \end{tabular}
	\quad
	\begin{tabular}{ lr }
	\toprule
	Preposition & Frequency \\
	\midrule
	PP $\rightarrow$ IN NP & 81\% \\
	PP $\rightarrow$ TO NP & 9\% \\
	\bottomrule
    \end{tabular}
\end{table}

Earlier works like OpenIE and ReVerb have looked at the extraction of subject-verb-object (SVO) triples. They were however based primarily on the availability of POS and chunker data. Structure of the verb and noun phrases were determined using statistical distribution of the phrases in text data. ClausIE used a dependency parser in resolving the SVO relations. 

Projects like DBpedia~\cite{auer2007dbpedia} were designed to extract structured data in the information box and map it to an ontology. Tgrep2~\cite{rohde2001tgrep2} enable us to extract and parse a tree without explicitly coding the rules. A set of regular expressions are used to extract matching sub-trees.

\begin{figure}[!h]
    \centering
	\caption{Number of correct non-redundant extractions.}
	\includegraphics[width=0.6\textwidth]{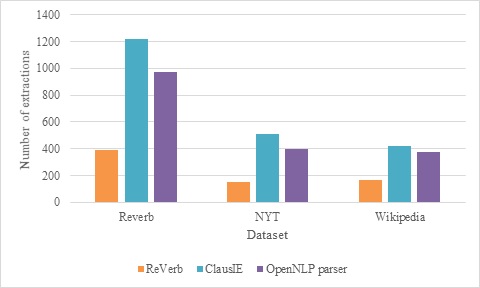}
	\label{fig:extractions}
\end{figure}

We evaluated a number of extractions on the ReVerb, Wikipedia and NYT dataset. We obtained the sample dataset from the ClausIE sources. We were able to extract more than 2000 SVO in the dataset with 1000 matching the ClausIE extractions. 

As all the extracted results are semantically accurate, the precision of the results is ~0.9. This value is independent of the dataset and is derived from the extraction grammar rules. The extractions are based on a rule based system and capture the syntax of the English language. Some of the SVO outputs are incorrect due to the ambiguities in the language parse tree including conjunctions in noun phrases. We verified the extracted triples to measure the recall of the data. The recall value is a function of the grammar. We can refine the rules to find additional triples in the data. This would increase the recall on the extracted values. We measured an average recall value of 60\% on the triples - Table 2. We used the extractions-all-labeled as a baseline for our computation. These include all the extractions from ReVerb, ClausIE and other OIE utilities.

We estimated a precision of 0.8 for the parser extractions. We found that the parser was able to extract 2-2.5 times the correct extractions of ReVerb and 80\% of the correct non-redundant ClausIE extractions - Fig.~\ref{fig:extractions}.

\begin{table}
\centering
\begin{tabular}{ lrr }
	\toprule
	& Precision & Recall \\
	\midrule
	NYT	& 0.8 & 0.64 \\
	Wikipedia & 0.8 & 0.71 \\
	ReVerb & 0.8 & 0.53 \\
	\bottomrule
\end{tabular}
\caption{Precision and recall values for various datasets}
\end{table}

\section{Conclusion}

We presented a methodology for extraction of subject-predicate-object triples in a text corpus. We plan to extend this work to a larger ontological engineering for knowledge inference. We found that a syntactic parser was able to accurately extract triples in a text. We explored opportunities to further extend this work in translating an unstructured corpus of data into a semantic ontology. A user is able to explore the text using a triples structure.

\bibliography{references}
\bibliographystyle{abbrv}

\section{Bibliography of Authors}

\begin{wrapfigure}{l}{0.2\textwidth}
  \begin{center}
    \includegraphics[width=0.2\textwidth]{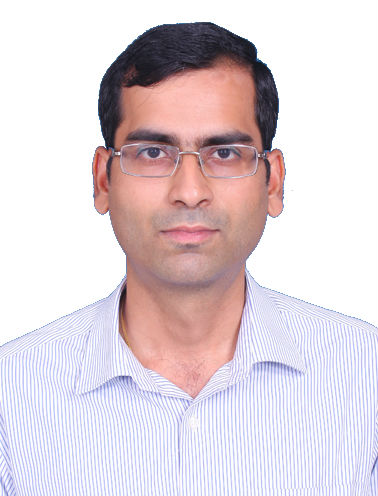}
  \end{center}
\end{wrapfigure}

Shaun D'Souza obtained a M.S.E. degree in Electrical Engineering from the University of Michigan, Ann Arbor and a B.S. degree in Computer Science, Electrical and Computer Engineering from Cornell University. He is currently working as a Technical Lead in the CTO Office at Wipro. His research interests include machine learning, compilers, algorithms and systems.

\end{document}